\documentclass[10pt,twocolumn,letterpaper]{article}

\usepackage{wacv}
\usepackage{times}
\usepackage{epsfig}
\usepackage{graphicx}
\usepackage{amsmath}
\usepackage{amssymb}
\usepackage{amsfonts}
\usepackage[ruled, lined, linesnumbered, commentsnumbered, longend]{algorithm2e}
\usepackage{xcolor}

\SetCommentSty{mycommfont}

\usepackage{graphicx}
\usepackage{multirow}
\usepackage{tabularx} % for stretchable tables
\usepackage{caption}
\usepackage{xspace}
\usepackage{booktabs}
\usepackage{subfigure}
\usepackage{amsmath,amssymb} % define this before the line numbering.
\usepackage{enumitem,xcolor}

\def \ie {{i.e.}\xspace}
\def \eg {{e.g.}\xspace}

\def \etal {{et al.}\xspace}

\newcommand{\miaojing}[1]{#1}

\newcommand{\para}[1]{\noindent \textbf{#1}}

\usepackage{amsmath,epsfig}
\usepackage{paralist} 
\usepackage[normalem]{ulem}
\usepackage{colortbl}
\def\wacvPaperID{1152} % Enter the WACV Paper ID here

\wacvfinalcopy % *** Uncomment this line for the final submission

\ifwacvfinal
\fi

\ifwacvfinal
\usepackage[breaklinks=true,bookmarks=false]{hyperref}
\else
\usepackage[pagebackref=true,breaklinks=true,colorlinks,bookmarks=false]{hyperref}
\fi

\ifwacvfinal
\else
\fi

\begin{document}

%%%%%%%%% TITLE
\title{Detecting Human-Object Interaction with Mixed Supervision}

 \author{Suresh Kirthi Kumaraswamy\\
 Univ Le Mans, CNRS, IRISA\\
{\tt\small kirthifame@gmail.com}
 \and
 Miaojing Shi\\
 King's College London\\
 {\tt\small miaojing.shi@kcl.ac.uk}
 \and
  Ewa Kijak\\
 Univ Rennes, Inria, CNRS, IRISA\\
 {\tt\small ewa.kijak@irisa.fr}
 }

\maketitle
\thispagestyle{empty}
\begin{abstract}
Human object interaction (HOI) detection is an important task in image understanding and reasoning.
%which bridges computer vision and language. The detection 
It is in a form of HOI triplet ${\langle human, verb, object \rangle}$, requiring bounding boxes for human and object, and action between them for the task completion. In other words, this task requires strong supervision for training that is however hard to procure. A natural solution to overcome this is to pursue weakly-supervised learning, where we only know the presence of certain HOI triplets in images but their exact location is unknown.  Most weakly-supervised learning methods do not make provision for leveraging data with strong supervision, when they are available; and indeed a naive combination of this two paradigms in HOI detection fails to make contributions to each other. In this regard we propose a mixed-supervised HOI detection pipeline: thanks to a specific design of momentum-independent learning that learns seamlessly across these two types of supervision.
%We propose a single network architecture which can accomplish weak and strong supervision separately and as well as jointly (mixed learning). 
%We realize that unlike separate weak/strong supervised learning, the mixed learning leads to adversarial situations during learning. 
%We elaborate on the problems with extensive ablation studies. We propose ways to overcome these situations through independent momentum based updates and ideas from curriculum learning for presenting the training samples during mixed learning. We also show usage of $mixup$ for regions, and region-pair proposals in the HOI detection, which is a little more involved than the mixup in the visual recognition cases.
Moreover, in light of the annotation insufficiency in mixed supervision, we introduce an HOI element swapping technique to synthesize diverse and hard negatives across images and improve the robustness of the model. Our method is evaluated on the challenging HICO-DET dataset. It performs close to or even better than many fully-supervised methods by using a mixed amount of strong and weak annotations; furthermore, it outperforms representative state of the art weakly- and fully-supervised methods under the same supervision.    

%\keywords{Weakly supervised learning, mixed learning, Human object interaction, HOI, momentum, Adam}
\end{abstract}

%\keywords{human object interaction, mixed-supervision, momentum-independent training, HOI element-swapping}

\section{Introduction}
%Deep learning methods have revolutionized the field of computer vision with their exemplary performances on recognition and detection tasks.

%Visual recognition~\cite{xx} has witnessed remarkable achievement thanks to the usage of deep neural networks (DNN). Human-object interaction (HOI) detection goes beyond the realm of visual recognition, it is related to visual relationship detection~\cite{}, but presents different challenges, as 
\begin{figure}
    \centering
    \includegraphics[scale=0.46]{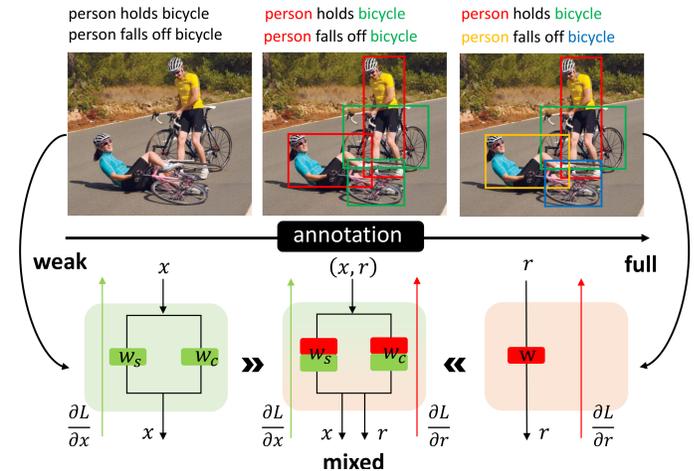}
    \caption{\small Human-object interaction detection with different levels of supervision. Top: annotation cost increases from image-level ($x$) labels (left) in weakly-supervised learning to region-level ($r$) bounding boxes (middle) and their correspondences (right) in fully-supervised learning. Bottom: our proposed mixed-supervised HOI detection pipeline (middle) enables joint learning of weakly- and fully-supervised HOI detection (left and right).      
    }\label{fig:introduction}
    \vspace{-1mm}
\end{figure}

%\begin{figure}
%    \centering
%    \includegraphics[scale=0.46]{figures/plot_wMIL_wo_MIL.png}
%    \caption{\small Plots for Mixed-Learning with and without Momentum Independent Learning (MIL). Without-MIL curve (\textit{blue}) is for a single configuration FS/WS = $30\perc / 70\perc$. Curves for With-MIL keeping FS fixed @$30\%$ and varying WS ($orange$), and keeping WS fixed @$30\%$ and varying FS ($green$).
%    }\label{fig:MIL_plot}
%\end{figure}

The task of human-object interaction (HOI) detection is defined as a detection of a triplet ${\langle human, verb, object \rangle}$, identifying not only the bounding boxes of human and object but also their interaction~\cite{gkioxari2018detecting,xu2019mm,gupta2019iccv,qi2018eccv,chao2018wacv,zhou2019iccv,delaitre2011nips,desai2010cvprw}. It is derived from visual relationship detection (VRD) of triplet  ${\langle object_1, predicate, object_2 \rangle}$
~\cite{lu2016eccv,gupta2015arxiv,gupta2008eccv,gupta2009pami,ramanathan2015cvpr,atzmon2016arxiv,zheng2019mm,sun2019mm,zhou2019mm}, but present different challenges: the predicates in VRD can be verbs (\eg ``push"), spatial (\eg ``on top of"), prepositions (\eg ``with"), comparative (\eg ``
taller than"), while in HOI they are mainly verbs. On the other hand, \emph{human-centric} interactions are more diverse and complicated, one person can easily interact with multiple objects in the meantime, \eg ``person wearing a jacket and riding a bicycle". This makes HOI a much more fine-grained task than VRD.

Intensive attention has been drawn to HOI alongside the introduction of new benchmarks, \ie V-COCO~\cite{gupta2015arxiv},  HICO-DET~\cite{chao2015iccv}, featured with diverse and numerous human-object interactions. For instance, in HICO-DET, there exist 600 HOI classes and 80 common object classes in total. Despite that recent advances have reported significant improvement~\cite{gupta2019iccv,qi2018eccv,chao2018wacv,zhou2019iccv}, the annotation cost is exponentially increased in these datasets. Given $N$ humans and $M$ objects in an image, the maximum number of HOI is $N \times M$, where we have to scan each of them and provide instance-level annotations (\eg bounding boxes) for the real ones (see Fig.~\ref{fig:introduction}: top-right). To alleviate this manual labor, we could provide only image-level HOI labels: a set of images are known to contain triplets of a certain HOI class, but the location and correspondence of objects are unknown in images (see Fig.~\ref{fig:introduction}: top-left); note the location of objects are assumed known sometimes (top-middle). Both cases can be conceptualized as weakly-supervised HOI detection.

There are few works that learn interactions from weak supervision. One representative is for VRD~\cite{zhang2017iccv}, which designs a weakly-supervised predicate prediction module inspired from the two-branch parallel structure in~\cite{bilen2016cvpr}. This can be easily adapted to HOI detection, as shown in Fig.~\ref{fig:introduction}: bottom-left. 
Nevertheless, in real application, instead of having only one type of labels, we can have a mixture of them: fully-labeled (instance-level), weakly-labeled (image-level), and even unlabeled. 

A generalized HOI detection framework for mixed supervision thus becomes necessary. 
An intuitive solution would be merging the weakly-supervised and fully-supervised HOI detection as a multi-task job,
%In this paper, we propose a unified framework that merges the weakly-supervised and fully-supervised HOI detection as a multi-task job, which is intuitively believed to be beneficial to each other. Such a solution is 
which is nevertheless not straightforward: the different quality of annotations between weakly-labeled and fully-labeled data, as well as their imbalanced ratios 
%between weak and full supervision 
should be considered. 
%This is illustrated in Fig.~\ref{fig:MIL_plot}: 
Fig.~\ref{fig:MIL_plot} illustrates an example: in the HICO-DET dataset, when adding different amounts of fully-labeled data (in red), results are either only slightly better than or even worse than learning with only weakly-labeled data (in grey). This simple combination, at best, does not exploit the full potential that could be derived from fully-supervised data, at worst, decreases the results obtained with weak supervision.
%learning with only weakly-labeled data (in grey), or even worse, depending on the amount of extra full supervision.
%The straightforward way is to merge the weakly-supervised and fully-supervised HOI detection as a multi-task job, which is intuitively believed to be beneficial to each other.
%However it is not realistic: %First, we need a stronger  weakly-supervised HOI pipeline to pair with the SOTA fully supervised pipeline so that they can both make contributions. Second, 
%Indeed,}% the optimization in the two learning manners is different as one focuses on the image-level and the other on the region-level ($x$ and $r$ in Fig.~\ref{fig:introduction}); the error surface toggles between the gradient from weak supervision and full supervision across different batches in the network. 
%a naive combinatorial loss actually leads to adversarial effect on both sides (see Sec.~\ref{sec:ablation}). 
This is the first challenge that needs to be solved in the mixed-supervised setting. Furthermore, HOI detection is a fine-grained task requiring the classification of similar interactions such as "eating", "drinking", "blowing". To be able to accurately distinguish them, diverse and hard negatives from similar interactions are essential for the network training. Nevertheless, due to the reason that many samples are weakly-labeled, interactions within them can not be clearly discriminated on the region-level; plus, some interaction classes are not even sufficiently collected. This poses another challenge for HOI detection. 
%is the inadequacy of labeled samples, which might be due to the natural long-tail distribution in many real world problem: some HOIs are common, \eg~ ``person riding bicycle", some are rare, \eg~``person riding tiger"; it could also be the subjective reason that only a few samples are elaborately labeled in the mixed supervision. 

%\vspace{0.2cm}
\para{Contributions.} We for the first time propose a generalized framework for mixed-supervised HOI detection (MX-HOI): % which addresses the aforementioned challenges. 
\begin{compactitem}
%\begin{itemize}[topsep=2pt,parsep=2pt,partopsep=2pt]

\item We integrate two state-of-the-art pipelines \cite{gupta2019iccv} and \cite{zhang2017iccv} for fully- and weakly-supervised HOI detection into a mixed-supervised pipeline.

\item To tackle the multi-task optimization in the mixed pipeline, we introduce a {momentum-independent} learning strategy to tackle the adversarial effect between full and weak supervision, by separating their gradient history in momentum learning. 

\item To tackle the annotation insufficiency in the mixed supervision, we introduce an HOI element swapping strategy to specifically harvest hard negatives across images for the weakly-labeled data. 
\end{compactitem}
%\end{itemize}
%We improve the WSPP module in~\cite{zhang2017iccv} with instance-level HOI label updating, making it a stronger weakly-supervised pipeline to pair with~\cite{gupta2019iccv}. Second, 
% Overall, we introduce a generalized HOI detection framework for the sake of mixed supervision.
\noindent By conducting our generalized HOI detection framework on the most challenging HICO-DET dataset, we show our method enables HOI detection with a mixed amount of supervision, \eg with 30\% fully labeled data and 70\% weakly-labeled data, we are able to retain 93.3\% accuracy of the setting of 100\% fully-labeled data. Furthermore, 1) our model improves both the state of the art weakly- and fully-supervised HOI detection methods~\cite{zhang2017iccv,gupta2019iccv} under the same supervision; 2) unlabeled data can also be leveraged into our pipeline following a "pseudo label" solution~\cite{lee2013icmlw}, where we can use the network trained on labeled data to infer labels of HOI pairs on unlabeled data.
\section{Related Work}
\para{Visual relationships} were originally used to help improve object localization~\cite{gupta2008eccv,kumar2010cvpr,sadeghi2011cvpr}, action
recognition and pose estimation~\cite{desai2012eccv,ramanathan2015cvpr} or semantic segmentation~\cite{gould2008ijcv}. Relationships that are often modelled between objects include verbs, actions, spatial and prepositions~\cite{sadeghi2011cvpr,yatskar2016cvpr,gupta2015arxiv,gupta2008eccv,gupta2009pami,ramanathan2015cvpr,atzmon2016arxiv,han2018mm,cui2018mm}.  
\cite{lu2016eccv} was the first work
to formulate the detection of visual relationships as a separate task. They propose to leverage
language priors from semantic word embedding to finetune the likelihood of a predicted
relationship. Subsequently, many researchers improved and generalized this model~\cite{yu2017iccv,liang2017cvpr,cui2018mm}. 
%In [51] the authors use
%external sources of linguistic knowledge and the distillation technique [12] in order to improve
%modeling performance, while [21] formulates the task of detecting visual relationships as a
%reinforcement learning problem. 
%The work [20] adopts 
Triplet metric learning is also adopted to optimize the visual feature connections among semantically related objects in~\cite{li2017cvpr,sun2019mm}. 
Attention~\cite{han2018mm,zheng2019mm} and spatial locations~\cite{zhou2019mm} are some other additive cues to visual relationship detection.  
%datasets ......
%The very recent
%work [17] investigates visual attention to detect the relationships.

%\vspace{0.2cm}

\para{Human-Object Interaction} is a concept related to visual relationship. The interactions between humans and objects are mainly focused on verbs, and are much more fine grained (\eg holding, hitting, throwing, touching) than relationships between generic objects. The study of HOI dates back to~\cite{delaitre2011nips,desai2010cvprw,gupta2009pami,prest2011pami,xu2019mm}, when most works were tested on small datasets. This issue was addressed by Chao~\etal~\cite{chao2015iccv} where they introduced a large dataset (HICO-DET) covering 80 common object categories and 600 HOI categories in total. Many recent works report their performance on this dataset and significant improvement has been achieved~\cite{qi2018eccv,chao2018wacv,gupta2019iccv,liuamplifying, gao2020drg,li2020pastanet,wan2019pose,wang2020learning,li2019transferable, ulutan2020vsgnet}. For instance, Qi~\etal~\cite{qi2018eccv} proposed a graph parsing neural network for HOI and was later extended by~\cite{hu2019iccv,xu2019cvpr,zhou2019iccv}; Gupta~\etal~\cite{gupta2019iccv} showed that a simple factorised model with appearance and layout encoding constructed from pretrained object detectors outperforms more sophisticated approaches; \miaojing{additional cues such as language features~\cite{gao2020drg}, parts based features~\cite{li2020pastanet, wan2019pose} are also exploited.} Our MX-HOI is built on the recent advance of \cite{gupta2019iccv}. 

%additional interaction learning~\cite{} and .
%{\color{blue} HOI detection is becoming sophisticated through the use of  } %Gupta~\etal~\cite{gupta2019iccv} showed that a simple factorised model with appearance and layout encoding constructed from pretrained object detectors outperforms more sophisticated approaches; we show the strength of MX-HOI by building on this simple model}.

\begin{figure*}[t]
    \centering
    \includegraphics[scale = 0.55]{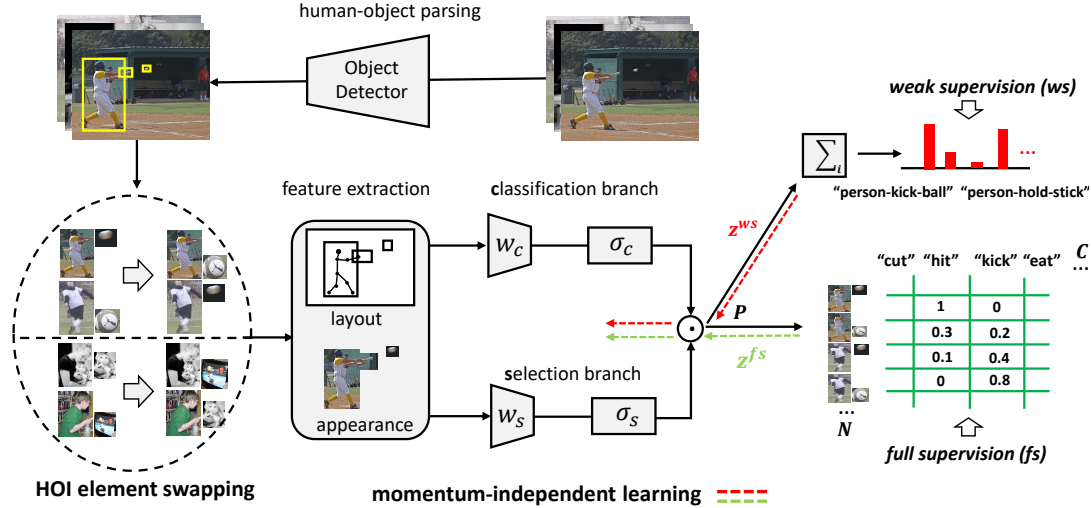}
    \caption{\small Illustration of our mixed-supervised HOI detection pipeline (MX-HOI). Human and object bounding boxes are obtained via an object detector. Human-object pairs are randomly created within an image and also across another image via the proposed HOI element swapping. HOI detection is realized via a two-branch structure in parallel for interaction classification and selection. Each branch consists of FC layer ($w_c$/$w_s$) for score prediction and softmax layer ($\sigma_c$/$\sigma_s$) for score normalization over rows or columns of the score matrix, respectively. The score matrices in the two branches are of size $N$ (human-object pairs) and $C$ (HOI classes) and are multiplied to produce the final matrix $P$.  Training data with full and weak supervision ($fs$, $ws$) are optimized with region-level and image-level ground truth, respectively. We introduce a momentum-independent strategy to enable the mixed-supervised learning with two momentum $z^{fs}$ and $z^{ws}$. Human-object pairs from two images are optimized in one batch. 
    }
   % \fix{Use the symbol $\odot$ for element-wise multiplication instead of $\otimes$ \cite{bilen2016cvpr}}
    \label{fig:method_overview}
\vspace{-1mm}
\end{figure*}

%\vspace{0.2cm}
\para{Weakly-supervised learning} in visual recognition is mostly focused on object detection~\cite{bilen2016cvpr,shi2016eccv,tang2017cvpr,shi2017iccv,tang2018pami,yang2019arxiv}. One seminal work in weakly-supervised object detection (WSOD) is~\cite{bilen2016cvpr}, where they designed a popular two-branch parallel structure followed by~\cite{tang2017cvpr,tang2018pami,yang2019arxiv}. Weakly-supervised relationship detection is more complex than WSOD as we need to detect individual objects for specific relations. Pretrained object detectors are normally assumed in this scenario~\cite{peyre2017iccv,zhang2017iccv,peyre2019iccv}. 
Peyre~\etal~\cite{peyre2017iccv} proposed a weakly-supervised discriminative clustering model to learn relations with only image-level labels; later on, they developed another model for transfer by analogy to obtain visual phrases of never seen relations~\cite{peyre2019iccv}. 
Zhang~\etal~\cite{zhang2017iccv} adopted the WSOD module in~\cite{bilen2016cvpr} to do weakly-supervised relationship detection and achieved very good results. %~\cite{peyre2017iccv,zhang2017iccv,peyre2019iccv} address the general relationship detection problem. There also exists weakly-supervised work solving the HOI problem~\cite{prest2011pami}. It learns the probability distribution of human-object spatial relations and train a binary classifier for action recognition.  This method involves hand-crafted features and probabilistic models, which would not be straightforward to extend to a deep learning framework.   
%\ewa{difference/limitation. Also comment on %~\cite{peyre2019iccv}} 
%We adapt~\cite{zhang2017iccv} to HOI task by simplifying it as part of our MX-HOI.We instead follow the recent work of~\cite{zhang2017iccv} 
In this work, we also adopt the WSOD module following~\cite{zhang2017iccv} and adapt it to be part of our MX-HOI pipeline.     

%\vspace{0.2cm}
\para{Mixed-supervised learning} normally refers to learning from a mixture of strongly labeled data and weakly-labeled data. For instance, Cinbis~\etal~\cite{cinbis2016pami} considered mixed supervision in object detection where some images are annotated with bounding boxes while some are only with image-level labels. Papandreou~\etal~\cite{papandreou2015iccv} studied the problem for semantic image segmentation from a combination of few strongly labeled (pixel-level annotations) and many weakly labeled (image-level labels or bounding boxes) images. Mixed-supervised learning can also be realized as leveraging an existing dataset of fully-labeled training images of non-target classes during the weakly-supervised learning of a new object category, which is connected to transfer learning, see \eg \cite{deslaers2012pami,shi2015bmvc,shi2017iccv,yang2020nips}.

\section{Mixed-supervised HOI detection}\label{sec:method}

%\ewa{Eliminate redundancies between into, sota,preliminaru and method overview.}

\subsection{Preliminary}\label{sec:preliminary}
%\para{Review of~\cite{gupta2019iccv} and~\cite{zhang2017iccv}. } 
We build our MX-HOI framework on two state of the art HOI works with full supervision~\cite{gupta2019iccv} and weak supervision~\cite{zhang2017iccv}, respectively.
\cite{gupta2019iccv} introduces a no-frills model for HOI detection where they use appearance features from pretrained object detectors, spatial features through box layout, and encode human pose keypoints, as shown in Fig.~\ref{fig:method_overview}: feature extraction. This is a no-frills detection without relying on attention or graph-based message passing~\cite{zheng2019mm,qi2018eccv}. It uses a factorized multi-layer perceptrons (MLPs) and introduces several new training techniques to improve the MLPs: eliminating a train-inference mismatch, rejecting easy negatives using indicator terms, and training with large negative to positive ratios. 

%While \cite{gupta2019iccv} relies on full supervision, 
\cite{zhang2017iccv} adopts the weakly-supervised object detection pipeline~\cite{bilen2016cvpr} for weakly-supervised predicate prediction (WSPP): it is accomplished via the {element-wise} multiplication of the predicate selection and classification branch (see Fig.~\ref{fig:method_overview}). The predicate score is softmax normalized over all candidate human-object box pairs with respect to a predicate class for the selection branch, and over all possible predicate classes with respect to one human-object pair for the classification branch, respectively. The predicate score in~\cite{zhang2017iccv} is obtained from a position-role sensitive score maps with a pairwise ROI pooling. To integrate it into the no-frills model above, we use the conventional ROI pooling. Predicate scores are predicted from the FC layers of the two branches, which is similar to the original structure in~\cite{bilen2016cvpr}. 

%To make the adapted WSPP stronger, we further introduce an instance-level HOI updating scheme. The instance-level HOI labels inferred from weak supervision act as pseudo ground truth for self-supervision~\cite{lee2013icmlw}. The pseudo ground truth will be updated if the predicted labels are above some thresh. 

\subsection{Overview}\label{sec:overview}
We introduce a mixed-supervised HOI detection framework (MX-HOI) as shown in Fig.~\ref{fig:method_overview}: the input for MX-HOI is region proposals output from a pretrained object detector. We follow the same procedure as in~\cite{gupta2019iccv} to extract both appearance and layout features for the human and object bounding boxes in a pair. Given human-object pairs, their region features are fed into the adapted two-branch predicate prediction structure from~\cite{zhang2017iccv} (Sec.~\ref{sec:preliminary}). The output of the two branches (matrices) are multiplied element-wise to produce one $N \times C$ matrix $P$ over $N$ human-object pairs (in a batch) and $C$ interaction classes. Each element $p_{ij}$ indicates the probability of the $i^\text{th}$ human-object pair having $j^\text{th}$ interaction type. For fully-labeled data, the predicate prediction is optimized on the region-level on matrix $P$, where a corresponding ground truth matrix is associated with each element being 1 or 0 indicating the true or false for the human-object interaction. For weakly-labeled data, the predicate optimization is on the image-level:  $P$ is accumulated over rows ($\sum_i p_{ij}$) to produce a $C$-dimensional vector where each element signifies the probability of the image containing certain HOI class. Similarly, an image-level ground truth vector with elements 1 or 0 is associated.  

This is a multi-task optimization defined jointly with full and weak supervision. The learning is not straightforward: the optimization in the two learning manners is different as one focuses on the image-level and the other on the region-level ($x$ and $r$ in Fig.~\ref{fig:introduction}); the error surface toggles between the gradients from weak supervision and full supervision across different batches in the network. %The learning is not straightforward, as the gradient flows on different surfaces (region-level and image-level) and they are functionally different for the two branches. 
We therefore propose a momentum-independent learning strategy (Sec.~\ref{sec:moment}). Besides, for the weakly-labeled data in mixed-supervised learning, we introduce an HOI element swapping strategy (Sec.~\ref{sec:mixproposals}) to further augment the hard negatives. 
%Unlabeled data can also be added into the framework using the network prediction as pseudo labels (Sec.~\ref{sec:unlabeled}). 
Loss function is given in the end (Sec.~\ref{sec:loss}). 

\subsection{Momentum-independent Learning}\label{sec:moment}
 
In the context of mixed-supervised learning, network weights are updated by either weak or full supervision, depending on the samples within the mini-batch. 
%The weight updating in a gradient descent based optimizer is given by, 
%\begin{equation}
%    w_t = w_{t-1} + \alpha\nabla f(w_{t-1})
%\end{equation}
%where $\nabla f(w_{t-1})$ is the current gradient, $w_t$ is the weight at iteration $t$, and $\alpha$ is the step size.
Most recent gradient descent based optimizers use momentum-based weights update. Let $w_t$ and $\nabla f(w_{t})$ be respectively the weight and the gradient at iteration $t$, and $\alpha$ be the step size, the momentum-based update rule is given by:
%\begin{subequations}
%\begin{equation}
\begin{align}  \label{eq:mom_based}  
\vspace{-3mm}
    w_t = w_{t-1} - z_t; ~~
    z_t = \beta z_{t-1} + \alpha \nabla f(w_{t-1})
    \vspace{-3mm}
    \end{align}
%\end{equation}
%\end{subequations}
where $\beta$ is the momentum parameter (usually $\beta \geq 0.9$) and $z_t$ is the momentum, which is dependent on all the previous gradient values.

Using momentum-based gradient descent can however be a problem in the mixed-supervised learning. In the fully-supervised case, the ground truth is directly given on the instance level such that the gradient of the loss function will accordingly backpropagate to the specific regions of the human and object in a pair. In the weakly supervised case, the ground truth is instead only given on the image-level, and predictions on all possible human-object pairs are aggregated together to the image-level for loss computation; at the backpropagation time, the gradient is distributed among all the human-object pairs. As a result, the gradients for full and weak supervision are computed on different error surfaces and are not compatible. Using one momentum to record both will make the network weight optimization toggles between the two sources of gradients across mini-batches. This indeed leads to an adversarial effect of the mixed-supervised learning (see the ablation study in Sec.~\ref{sec:ablation}). 

To mitigate this, we propose to bootstrap the mixed-supervised learning with two independent momentum $z_t^{ws}$ and $z_t^{fs}$ to record the gradient history of weak and full supervision separately. $z_t^{ws}$ will be used and updated only with weakly-labeled samples in the mini-batch, while $z_t^{fs}$ will be instead used for the fully-labeled samples. $w_t$ however is remained to be shared in the network such that the weakly- and fully-supervised pipeline are jointly optimized: 
\begin{equation}\label{eq:mom_modified}
\begin{split}
\vspace{-5mm}
   w_t = w_{t-1} - z_{t}^{ws} ; \;\;\;\; z_{t}^{ws} = \beta z_{{t-1}}^{ws} + \alpha \nabla f(w_{t-1}) \\
   w_t = w_{t-1} - z_{t}^{fs} ; \;\;\;\; z_{t}^{fs} = \beta z_{t-1}^{fs} + \alpha \nabla f(w_{t-1})
\end{split}
\end{equation}
%\end{subequations}

%where $z_{t}^{ws}$ and $z_{t}^{fs}$ are weak and full supervision momentum updates respectively. This is very simple to implement, in \textit{PyTorch} it is just declaring two optimizers, $optimizer_{ws}$ and $optimizer_{fs}$ and use corresponding $optimizer_{\times\times}.step()$.

%\miaojing{Suresh, could you finish this part ? in a way that first posing problems, them solutions, third performance xx}

\subsection{HOI element swapping}\label{sec:mixproposals}

\begin{figure}
    \centering
    \includegraphics[scale=0.5]{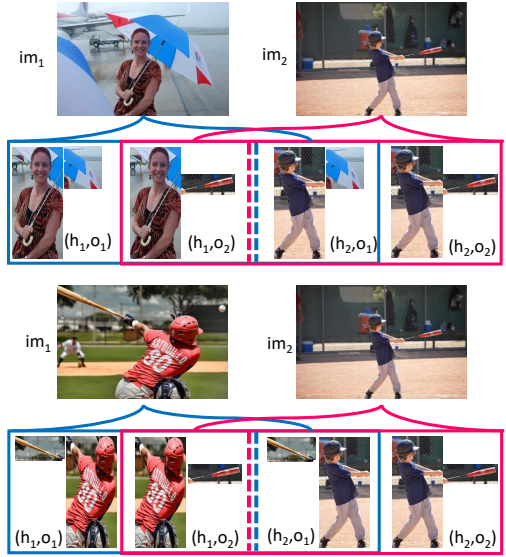}
    \caption{Illustration of HOI element swapping. Top: Object classes from the two images are different. Bottom: same object class in the two images. Swapped pairs ($h_i$, $o_j$) are negatives due to wrong object class (top) or wrong spatial layout (bottom).}
    \label{fig:Mixup_HO}
    \vspace{-5mm}
\end{figure}

HOI detection is a fine-grained task. To accurately classify similar interactions, diverse and hard negatives are needed. In the fully-supervised case, where region-level ground truth are available, this can be achieved via choosing the false positive class of large confidence score or false positive region of large intersection-over-union (IoU) with ground truth.
%\fix{as was addressed in~\cite{gupta2019iccv}}. \ewa{I don't see this process in the paper}  
While in the weakly-supervised case, where only image-level ground truths are available, conventional manners of finding negatives no longer apply. We instead introduce an HOI element swapping way to collect diverse and hard negatives across images.

%Conventional approaches tend to often lower the threshold from the RPN to produce more proposals, hoping to find hard negatives within them{~\cite{gupta2019iccv}}. This however is not quite efficient as many proposals with low confidence are not even good  detections of humans or objects. We instead introduce another way to collect hard negatives across images using existing proposals. 

Suppose that two images $im_1$ and $im_2$ contain one human $h$ and one object $o$ inside each, respectively.
The standard way to create the candidate human-object pairs is to group ($h_1$, $o_1$) and ($h_2$, $o_2$) within each image (see Fig.~\ref{fig:Mixup_HO}). To further augment negatives, a simple way is to lower the threshold from the RPN to produce more proposals; but this is inefficient as many proposals with low confidence scores are not good detection of humans or objects, and we do not have ground truth bounding boxes to distinguish them in the weakly-supervised setting. Hence, we propose to keep a fair detection of humans and objects within each image and augment the human-object pairs across images by swapping their HOI elements: given ($h_1$, $o_1$) from $im_1$ and ($h_2$, $o_2$) from $im_2$, we mix the human proposal from the $im_1$ with the object proposal from  $im_2$ and vice versa: ($h_1$, $o_2$) and ($h_2$, $o_1$); this gives us two more mixed human-object pairs.  
One image may contain more than one human or object. Considering there are $H_1$ humans and $O_1$ objects in $im_1$, $H_2$ humans and $O_2$ objects in $im_2$, selecting all humans and objects from the two images will produce $(H_1 + H_2) \times (O_1 + O_2)$ pairs in total for the two images. 
which is far too much. In practice, we remove those easy negatives with low confidence scores such that the number of human-object pairs is kept the same to that of the original number in two images, \ie  $H_1 \times O_1 + H_2 \times O2$.  
%randomly suppress those HOI pairs whose prediction scores are below some thresh in both fully- and weakly supervised pipeline.
%\miaojing{We then just associate one different object from $im_2$ to each human of $im_1$, creating $\min(H_1,O_2)$ new HOI pairs for one image $im_1$.}

By doing this, we can obtain more diverse combinations of HOI pairs, where many swapped HOI pairs coming from two images might look like positive HOI pairs playing the role of hard negatives. For instance, in Fig.~\ref{fig:Mixup_HO}: bottom, the object class from two images is the same in particular, yet the swapped human-object pairs  ($h_1$, $o_2$) and ($h_2$, $o_1$) are still negatives due to the wrong spatial layout.  
%To remove those easy negatives, we randomly suppress those HOI pairs whose prediction scores are below some thresh in both fully- and weakly supervised pipeline.

%In the fully supervised case, we can simply set ground truth labels of those mixed HOI pairs as none to all the classes, making them negatives. 
The augmented human-object pairs as shown in Fig.~\ref{fig:Mixup_HO} can be hard negatives for both images. To efficiently optimize the learning on two images, we propose to aggregate all the human-object pairs from two images to form one image-level  HOI label vector, where the corresponding ground truth is the HOI labels from both images. Apart from efficiency,  another benefit of doing this, compared to optimizing the image separately,  is that the positive human-object pairs from one image could also serve as negatives for the other image if they are of different HOI labels or as positives if they are of the same HOI label. 

\subsection{Loss function}\label{sec:loss}
Loss function is defined within each mini-batch depending on the supervision $\mathbb{S}$ of the input samples, which can be full supervision $(\mathbb{S = FS})$ or weak supervision $(\mathbb{S=WS})$:

\begin{equation}\label{eq:loss}
\begin{split}
{\mathcal{L}_{\text{mini-batch}}} & = \sum_{j=1}^{C} \bigg (\mathbf{1}(\mathbb{S = FS}) \frac{1}{N} \sum_{i=1}^{N} BCE(y_{ij},p_{ij})  \\
& + \mathbf{1}(\mathbb{S = WS}) BCE(y_j,p_j)\bigg ) \\
\end{split}
\end{equation}
%\mathcal{L}_{mini-batch}=  \sum_{l=1}^{C} \left(\mathbbm{1}(s=fs) \frac{1}{N} \sum_{j=1}^{N} BCE(y^j,p_l^j) + \mathbbm{1}(s=ws) BCE(y,p_l) + (\mathbbm{1}(s=us) \frac{1}{M} \sum_{j=1}^{M} BCE(y^j,p_l^j) \right)
BCE is the binary cross-entropy; $p_{ij}$ is the probability of $j^\text{th}$ HOI class for the $i^{th}$ human-object pair, where there are $N$ pairs and $C$ classes in total; $p_j$ is the probability of the $j^\text{th}$ HOI class for the given images in the mini-batch. The former is defined for the fully-labeled data with region-level ground truth, while the latter is defined for the weakly-labeled data with image-level ground truth only. In practice, we feed the features of human-object pairs from two images in each mini-batch. %The mini-batch size is normally 1, \miaojing{xx} . 
Referring to Sec.~\ref{sec:overview}, the ground truth for fully-labeled data is given in a form of a matrix, where each element $y_{ij}$ indicates whether the $i^{th}$ human-object pair with $j^\text{th}$ HOI class is true or false. $y_{ij} = 1$ if the human and object boxes in the $i^{th}$ pair have an IoU greater than 0.5 with a ground truth box-pair of the $j^\text{th}$ HOI class. The ground truth for the weakly-labeled data 
is given in a form of $C$-dimensional vector where its element $y_j = 1$ if the $j^\text{th}$ HOI class occurs in any of the two images.

%The sample can be one HOI triplet with full supervision or one image with weak supervision. 
%$N$ indicates the total HOI triplets in a fully-labeled image. $p_{ij*}$ is computed from (\ref{eq:unlabel}), where we use $M$ to denote the total number of HOI triplets assigned with pseudo labels in an image. 

%\miaojing{Explain : where the loss is computed on the architecture depending on full/weak supervision (and consequently from where the gradient backpropagation operate)}

%HOI detection is process of triplet, $<Human, Object, Predicate>$ detection. There are several approaches with CNN followed by graph based inferencing and fully CNN based. We follow fully CNN based as it is simple and easy to elucidate.
%We follows no-frills based HOI \cite{gupta2019iccv} where authors use pretrained object detection and pose estimation frame work outputs as the features. Here 

\section{Experiments}
\label{sec:experiments}
\subsection{Experimental setup}\label{Sec:experimentsetup}
\para{Datasets.} 
Ever since its introduction in \cite{chao2018wacv}, the HICO-DET has become the defacto standard dataset for human object interaction detection. The dataset has a total of 47,776 images: 38,118 (80\%) are used for training and 9,658 (20\%) for testing. 
%We follow~\cite{chao2018wacv} to use an 80-20 split of the training images to generate the actual training and validation sets. 
Each image is provided with the $\langle human, object, predicate\rangle$ triplets which include human and object bounding boxes and HOI classes. 
It covers 80 object categories and 117 interactions, which result into 600 HOI classes in total. These classes are subdivided into 138 \textit{rare} ones, whose training samples are less than $10$ images; $462$ \textit{non-rare} ones, whose training samples are more than $10$ images.
On average, 1.67 HOI triplets are annotated in each image.  HICO-DET is a much bigger dataset compared to the previous V-COCO dataset~\cite{gupta2015arxiv}.  In line with recent works on HOI detection~\cite{gupta2019iccv,bansal2020aaai}, we evaluate our method on the large-scale HICO-DET to offer comprehensive study and in-depth analysis on it.  

%The other HOI dataset we used is the V-COCO dataset \cite{gupta2015arxiv} that is build off the MS COCO dataset \cite{mscoco14}. It contains  a  total  of  10,346 images with 16,199 people instances. Each image has also annotations for 80 object categories along with segmentation mask for all objects, and 26 different actions for each annotated person. On average images have 1.57 people annotated with action labels per image, and people do on average 2.87 actions at the same time.  

%The work \cite{chao2018wacv} also proposes broad subdivision of the HICO-DET dataset with  
%In our implementation we further split the training data in to 80/20 for training and validation respectively.

\vspace{0.1cm}
%\para{Data splitting strategies for mixed supervision.}
%Specifically, we assume two settings in HICO-DET for the data split in mixed supervision which we refer to as \emph{image-split} and \emph{class-split}. 
\para{Data splitting.}%In the \emph{image-split} setting, we randomly split the training images
The training images are randomly split with different ratios of weakly- and fully-labeled data (denoted as WS and FS). The default WS/FS ratio is set to {70/30 and 30/70} where 70\% (30\%) data from the training set are weakly-labeled and the rest are fully-labeled. We also evaluate different WS/FS ratios ranging between 100/0 and 0/100 in the experiments. Some more settings regarding unlabeled data and class-split are also presented in the end. 
% \sout{To demonstrate the generalizability of our MX-HOI, we also 1) provide the result for the split WS/FS/US of 30/40/30 to allow for the inclusion of unsupervised data (US) in our pipeline; 2) evaluate our pipeline in a class-split setting by randomly split the whole HOI classes into 50\% classes with strong annotations and 50\% classes with  Images from the first 50\% classes are trained with full supervision, the second 50\% with weak supervision. }

%where 30\%, 40\% and 30\%  data are respectively weakly, fully and unsupervised, using a pseudo-label approach.            

%In \emph{image-split} setting, each class can contain both weakly- and fully-labeled samples. In contrast, in the \emph{class-split} setting, we randomly split the HOI classes with different ratios of weakly- and fully-labeled classes. Samples of each class can only be with one type of supervision, weak or full. We assume two default ratios for WS/FS as 70/30 and 30/70, where 70\% (30\%) classes from the training set are with weakly-labeled data, while the rest of classes are with fully-labeled data. The \emph{class-split} setting is indeed a transfer learning setting where we evaluate the cross class performance by transferring knowledge from classes with fully-labeled data (source) to classes with weakly-labeled data (target).  

\vspace{0.1cm}
\para{Implementation details and evaluation protocol.}
Following~\cite{chao2018wacv}, the human and object detection results are taken from the top scoring output of a Faster-RCNN pretrained on MS-COCO~\cite{mscoco14}. 
%The top scoring objects detected by Faster-RCNN are considered candidates for the HOI detection. 
Each human is paired with all the objects within an image. 
Faster-RCNN produces numerous candidate bounding boxes. For each object, we filter the 30-top performing boxes depending on the detection scores.
For fully-labeled data, ground truth HOI triplets are provided with human/object bounding boxes and their interaction. 
%A positive match from any human-object pair to a ground truth has to satisfy the IoU criteria (\eg IoU $>$ 0.5) for the detection of individual human and object, as well as  
%hence the human-object pairs are shortlisted via the IOU matched with the ground truth bounding boxes and corresponding elements of the label vector are labeled 0/1. 
For weakly-labeled data, only image-level HOI labels are provided meaning that the real correspondence from a human detection to an object is not given in the image. For unsupervised data, no HOI labels are provided. \miaojing{The network is trained with a mini-batch containing the set of region proposal pairs in two images, which are randomly selected from either the fully-labelled set or the weakly-labeled set. This is done once before the training for efficiency. Two images from the weakly-labeled set are applied with element swapping. }
%mini-batch size, number of positives versus negative in batch. with and without HES}. 
The learning rate is 1e-3 and 1e-4 for weakly- and fully-labeled data, respectively. We train 40,000 iterations in total. 
% {\color{blue}In case of mixed supervision learning data is given either FS or WS label corresponding to it's nature of supervision.}
%the above described human object proposals are retained but the 600 dimensional binary label vector is image level. We implement a single network to be used for weak, full and weak-full supervision learning. 

Evaluation of HOI detection employs the widely used mean average precision (mAP) metric where a prediction is considered correct only if its HOI class label is correct, and its human and object bounding boxes have an Intersection over Union (IoU) larger than $0.5$ with their respective ground truth bounding boxes. {

\subsection{Ablation study}\label{sec:ablation}

In this section, we first justfiy the importance of our proposed new elements MIL and HES in order to enable a meaningful mix-supervised HOI detection. Next, we give the result of MX-HOI generalizing over different WS/FS ratios from 0/100 to 100/0. 

\vspace{0.1cm}
\para{Using weak and strong annotations.}
To start with our ablation study, we first train our HOI detector with weak annotations only; next, we train the detector with both weak and strong annotations. We illustrate the mAP on the test set in Fig.~\ref{fig:MIL_plot}: by default 70\% (30\%) data are chosen as weakly-labeled and the rest are as fully-labeled. The results show that without using our proposed MIL (w/o MIL), adding FS data can perform even worse than using WS data only: for example using 70\% of the training data, all weakly-labeled (WS/FS = 70/0), it (\emph{weak only}, grey) yields a mAP of 14.68; when adding fully-labeled data (WS/FS = 70/30), the detector performs an even worse mAP 13.17 (red). %This shows that it is not straightforward to train the weakly- and fully-labeled data together, which indeed leads to an adversarial effect to each other. 
This illustrates well the adversarial effect between the weakly- and fully-labeled data.
In order to tackle the problem, we first tried a sequence training (ST) strategy~\cite{shi2016eccv,liu2019cvpr}, where all fully-labeled data (resp. all weakly-labeled data) are presented in the network with some epochs before the weakly-labeled data (resp. fully-labeled data) are added in. We denote the strategy as w/ ST-F when the fully-labeled data are trained first, or w/ ST-W when the weakly-labeled data are trained first.  The results are improved in this manner but not too much (see Fig~\ref{fig:MIL_plot}). Next, we introduce our momentum-independent learning strategy to specifically enable the mixed-supervised HOI detection.

\vspace{0.1cm}
\para{Momentum-independent learning (MIL).} %Referring to Sec.~\ref{sec:moment}, single momentum is originally used in the mixed-supervised learning as above, which results in the inconsistency of gradients in the network backpropagation. To tackle this, 
To tackle the inconsistency of gradients in the network backpropagation,
we introduce two independent momentum to record the gradient history of full and weak supervision separately as they are computed on different error surfaces and are functionally different in the network. This is conceptualized as momentum-independent learning and is a key element of our MX-HOI pipeline.  Having a look at Fig.~\ref{fig:MIL_plot}, ours (w/ MIL) significantly increases the mAP to \eg 16.63 for WS/FS = 70/30 and 17.47 for WS/FS = 30/70. This demonstrates the importance of our proposed MIL to enable a meaningful mixed-supervised HOI detection. Some more detailed comparisons between our MIL and other variants on the rare and non-rare classes are shown in Table~\ref{tab:MIL}. 

\vspace{0.1cm}

\begin{figure}
    \centering
    \includegraphics[scale=0.35]{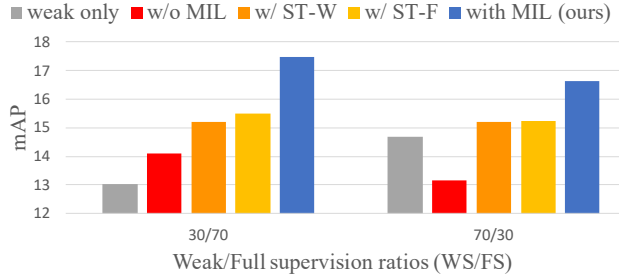}
    \vspace{-2mm}
    \caption{Using weak, strong and mixed annotations for HOI detection. MIL: momentum-independent learning. ST-W: sequence training with weakly-labeled data first; ST-F: sequence training with fully-labeled data first; 
    }\label{fig:MIL_plot}
\end{figure}

\begin{table}[t]
  	\setlength{\tabcolsep}{1.5pt}
    \centering
    \small
    \begin{tabular}{cccc|ccc} 
    \toprule

   \textbf{Method} &   \textbf{WS/FS} &  \textbf{Rare} & \textbf{Non-Rare} & \textbf{WS/FS} & \textbf{Rare} & \textbf{Non-Rare}\\
        \midrule
        weak only & 70/0  & 	11.84	&	15.72 & 30/0 	&	8.68	&	14.11 \\
        w/o MIL  &  70/30  & 		8.88	&	14.45 &  30/70 	&	9.17	&	15.74 \\ 
        w/ ST-F & 70/30 & 10.41	& 16.69	&  30/70  &	10.52 &	 17.00\\ 
        w/ ST-W  & 70/30  & 10.17	& 16.71& 30/70 	& 11.03	& 16.43	\\
        with MIL (ours) &  70/30 & 	\textbf{12.36}	&	\textbf{17.91} & 30/70 	&	\textbf{12.79}	&	\textbf{18.80}\\
\bottomrule
    \end{tabular}
   \vspace{-2mm}
 \caption{Ablation of momentum-independent learning (MIL) in MX-HOI on HICO-DET dataset. mAP is reported. }
    \label{tab:MIL}
\end{table}

%\begin{table}[ht]
%     \centering
% \begin{tabular}{|c|c|c|c|c|c|c|c|c|c|c|c|c|} \hline
%		\multicolumn{5}{|c|}{Num of \textit{Opti}s}	        &&				\multicolumn{7}{|c|}{Num of \textit{lr}s}													        \\\hline
%				\multicolumn{5}{|c|}{}	        &&				& \multicolumn{3}{|c|}{Single} & \multicolumn{3}{|c|}{Double}				\\ \hline

%\textit{Num opti.s} & \textit{WS/FS}	&		\textit{F}	                &	\textit{R}    &	\textit{NR} &&	%\textit{WS/FS}	&	\textit{F}	                &	\textit{R}    &	\textit{NR}	        &	\textit{F}	&	\textit{R}	&	\textit{NR}\\\hline
%Single	& 70/0	&		14.18	&	11.04	&	15.12	&&	50/50	&	14.69	&	9.47	&	16.24	&	16.18	&	11.18	&	17.67	\\\hline
%Single	& 70/30	&		13.17	&	8.88	&	14.45	&&	30/70	&	15.02	&	10.68	&	16.31	&	16.95	&	12.13	&	18.39	\\\hline
%Double	& 70/30	&		15.9	&	11.92	&	17.08	&&	\multicolumn{7}{|c|}{} \\\hline

%     \end{tabular}
%     \caption{\textbf{Need for two optimizers} and two lr.}
%     \label{tab:single_vs_double}
%\end{table}

\begin{figure*}[t]
\begin{center}
\includegraphics[scale=1.5]{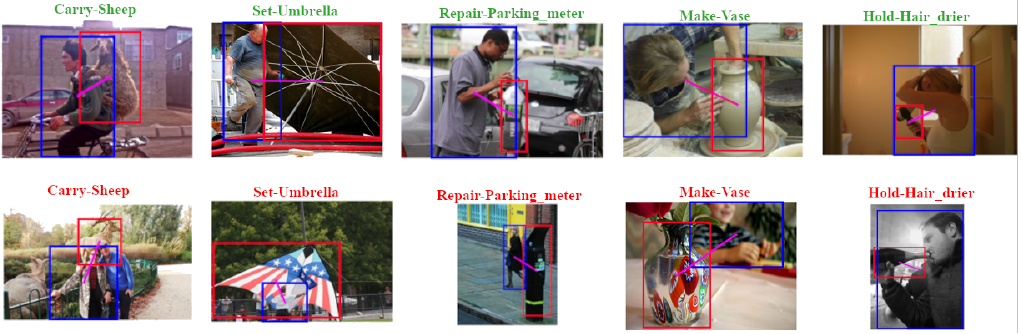}
%\includegraphics[scale=0.45]{./figures/hoi_image_corrected.png}
%\includegraphics[scale=0.5]{./figures/hoi_image_v3.pdf}
%\caption{Examples of detections corrected by HOI element swapping (HES).}
\vspace{-3mm}
\caption{Examples of correct detections (top) and incorrect detections (bottom) with MX-HOI. The classes shown here are Carry-Sheep, Set-Umbrella, Repair-Parking-meter, Make-Vase and Hold-Hairdryer.}
\label{fig:hes}
\vspace{-3mm}
\end{center}
\end{figure*}

\vspace{0.1cm}
\para{HOI element swapping (HES).} Referring to Sec.~\ref{sec:mixproposals}, HOI element swapping is introduced for hard negative harvest on weakly-labeled data. To verify its effectiveness, we ablate it in Table~~\ref{tab:HES} by comparing with MX-HOI without HES. We vary WS/FS from 100/0 to 30/70 and show that on different mixed levels, HES always helps the HOI detection. For instance, when WS/FS = 70/30, MX-HOI yields +0.81\% improvement over MX-HOI (w/o HES). 
%\fix{Examples of images correctly classified thanks to the HES are shown in Fig.~\ref{fig:hes}.}

Additionally, we also apply HES on fully-labeled data, \ie WS/FS = 0/100, and obtain the mAP 17.04, 13.35, 18.11 on full, rare and non-rare classes, respectively, which actually harms the performance on non-rare classes while helps a bit on rare classes comparing to 17.82, 12.91, and 19.17 in Table~\ref{tab:ratio1}. Rare classes do not have adequate training samples, HES can help provide hard negatives; while for non-rare classes, hard negatives can be directly mined via the given bounding box ground truth. Overall, we did not find HES to be effective for fully-labeled data in general.  
%as hard negatives can be directly mined via the given bounding box groundtruth.  

\begin{table}[t]
  	\setlength{\tabcolsep}{3.5pt}
    \centering
    \small
    \begin{tabular}{ccccc} 
    \toprule

  \textbf{Method}  &   \textbf{WS/FS} &  \textbf{Full} & \textbf{Rare} & \textbf{Non-Rare} \\
        \midrule
        ours (w/o HES) & 100/0  & 15.14 & 10.65  & 16.48 \\
        ours &  100/0  &\textbf{16.14} & \textbf{12.06} & \textbf{17.50}\\
        \midrule
          ours (w/o HES) & 70/30 & 15.82 & 10.39 & 17.41	\\
        ours &  70/30  & 	\textbf{16.63}	&	\textbf{12.36}	& \textbf{17.91}	\\ \midrule
         ours (w/o HES) & 30/70  &	16.73	&	12.00	    &	18.14	\\
        ours &  30/70   & \textbf{	17.47}	&	\textbf{12.79}	& \textbf{18.80}	\\ 
     \bottomrule
    \end{tabular}
    \vspace{-2mm}
 \caption{Ablation of HOI element swapping (HES) in MX-HOI on HICO-DET dataset. mAP is reported. }
    \label{tab:HES}
 \vspace{-2mm}
\end{table}

\vspace{0.2cm}
\para{WS/FS variations.} We offer the results of MX-HOI with different WS/FS: 100/0, 80/20, 70/30, 50/50, 30/70, 20/80 and 0/100 in Table~\ref{tab:ratio1}. One can see that the performance increases with an increase of FS for both rare and non-rare classes, as more fully-labeled data are added into the training. The overall full mAP increases from 16.14 to 17.82. In Table~\ref{tab:ratio1}, we also show the result of fixing either WS, or FS to 30\% while varying the other: the performance increases along with the training (sub-)set size;
%the performance increases either way when we add more FS/WS data in; 
but the improvement margin is bigger when fixing WS and increasing FS compared to fixing FS and increasing WS.
%{\color{blue}Effectiveness of FS over WS can also observed in non-MIL, refer to rows 2-4 of Table-1}. 
All these results make perfect sense for MX-HOI: adding more labeled data, regardless WS or FS, increases its performance; FS data in general provides more help than WS data.   
Examples of MX-HOI with WS/FS=70/30 are given in Fig.~\ref{fig:hes}.

\begin{table}[t]
  	\setlength{\tabcolsep}{3.5pt}
    \centering
    \small
    \begin{tabular}{cccc} 
    \toprule

        \textbf{Supervision (WS/FS)} &  \textbf{Full} & \textbf{Rare} & \textbf{Non-Rare} \\
        \midrule
                 100/0  &  16.14 & 12.06 & 17.50 \\
                 80/20 & 16.49 & 12.28 & 17.81 \\
              %  \textcolor{blue}{ 70/30}  & 15.82 & 10.39 & 17.41 \\
                 70/30  & 16.63 & 12.36 & 17.91 \\
                 50/50  & 17.08 & 12.58 & 18.17 \\ 
               %  \textcolor{green}{30/70}  & 16.73 & 11.48 & 18.14 \\ 
                30/70  & 17.47 & 12.79 & 18.80 \\ 
                 20/80 &  17.60 &   12.85  & 18.95  \\
                 0/100  & 17.82 & 12.91 & 19.17 \\ \bottomrule
                 30/30  &	16.05	&	11.64	&	17.37\\
                 30/50  & 	16.84	&	11.81	&	18.47\\ 
            %     \textcolor{green}{30/70}  &	16.73	&	12.00	    &	18.14 \\ 
                30/70  & 17.47 & 12.79 & 18.80 \\ 
                 50/30  &	16.34	&	12.04	&	17.69	\\ 
            %     \textcolor{blue}{70/30}  & 	15.55	&	10.79	&	16.97\\ 
                70/30  & 16.63 & 12.36 & 17.91 \\
            \bottomrule
    \end{tabular}
    \vspace{-2mm}
    \caption{Different ratios of WS/FS in MX-HOI on HICO-DET dataset (top). Fixing WS (resp. FS) ratio and varying the other (bottom). mAP is reported.}
    \label{tab:ratio1}
    \vspace{-3mm}
\end{table}

\subsection{Comparison to state of the art}

%\begin{figure*}[t]
%\begin{center}
%\includegraphics[scale=0.3]{./figures/white.png}
%\caption{Example of wrong detections with the SOTA No-Frills~\cite{gupta2019iccv} correctly classified with MX-HOI framework with full supervision. Improvements are due to the softmax layers introduced.}
%\label{fig:sota_comp}
%\end{center}
%\end{figure*}
In Table~\ref{tab:sota-res}, we first compare our method to its lower and upper bounds denoted respectively by WS-No-Frills and No-Frills in the following. WS-No-Frills is an adaption of a representative weakly-supervised relationship detection module~\cite{zhang2017iccv} onto the SOTA HOI detection pipeline~\cite{gupta2019iccv}, which produces mAP 15.14, 10.65, and 16.48 on full, rare and non-rare classes. Our MX-HOI under the same setting WS/FS=100/0 improves the result to 16.14, 12.06, 17.50 due to the adoption of HOI element swapping. In fully-supervised setting (WS/FS = 0/100), our MX-HOI also improves the No-Frills~\cite{gupta2019iccv} by +0.6\%, this is attributed to our two-branch softmax~\cite{bilen2016cvpr} in No-Frills \miaojing{(This softmax is applied in subbranches before the final classification head).} Despite MX-HOI is introduced for mix-supervised HOI detection, it also improves the SOTA bounds as side benefits. 
%\fix{Fig.~\ref{fig:sota_comp} gives examples of gives examples of  detections improved with MX-HOI compared to the SOTA No-Frills.} 

In the mixed-supervised setting, MX-HOI retains 93.3~\% accuracy of the SOTA No-Frills by using a mixture of 30\% fully-labeled data and 70\% weakly-labeled data. To compare with it, we implement a naive \emph{multi-stage training pipeline}: it first trains the model on 30\% fully-labeled data, then infers the HOI class probabilities on the human-object pairs in the rest 70\% weakly-labeled images; the HOI triplet with the largest probability is selected as pseudo ground truth for each given HOI label on the image-level in weakly-labeled data. These selected HOI triplets are mixed with existing fully-labeled data to train the network again. The network remains a fully-supervised pipeline in this manner. This process would repeat several rounds until the convergence of the model. We obtain mAP 15.23, 10.63, and 16.61 under the setting of WS/FS = 70/30, which is much lower than our MX-HOI (16.63, 12.36 and 17.91).

We also compare MX-HOI with other recent arts 
%{\color{blue} (sans language-embedding and graph interaction learning)}
~\cite{xu2019cvpr,gao2018bmvc,qi2018eccv,ulutan2020vsgnet,wan2019pose,li2019transferable} using 100\% supervision. One can see that with WS/FS = 70/30, MX-HOI performs very close to the SOTA. %Of course, some credits should be given to the strong pipeline introduced by~\cite{gupta2019iccv} and~\cite{bilen2016cvpr}. 
%\footnote{in arxiv version of \cite{gao2018bmvc}, reported mAP(\%) on three category sets for HICO-DET: 14.84, 10.45, 16.15.}.
%\footnote{Reported mAP(\%) in arxiv version of \cite{gao2018bmvc}: 14.84, 10.45, 16.15.}

\begin{table}[t]
    \centering
    \begin{tabular}{c|c|ccc} 
        \textbf{Methods} &\textbf{WS/FS} &  \textbf{Full} & \textbf{Rare} & \textbf{Non-rare} \\ \hline 
                WS-No-Frills &  \multirow{2}{*}{100/0} & 15.14 & 10.65  & 16.48 \\
                 \textbf{MX-HOI} & & \textbf{16.14} & \textbf{12.06} & \textbf{17.50} \\ 
                 \cline{1-5}
                  \textbf{MX-HOI} & 70/30  &  16.63 & 12.36 & 17.91 \\
                %  \textbf{MX-HOI} &50/50  & 16.18 & 11.18 & 17.67 \\ 
                   \textbf{MX-HOI} &30/70  & 17.47 & 12.79 & 18.80\\
                 \cline{1-5}
                 \textbf{MX-HOI} & \multirow{6}{*}{0/100}  & \cellcolor{gray!25}{17.82} & \cellcolor{gray!25}{12.91} &\cellcolor{gray!25}{19.17} \\
                 No-Frills \cite{gupta2019iccv} &  & 17.18 & 12.17 & 18.68\\
                {VSGNet} \cite{ulutan2020vsgnet} &  & \textbf{19.80} & \textbf{16.05} & \textbf{20.91}\\
               {PMFNet} \cite{wan2019pose} &  & 17.46 & 15.65 & 18.00\\
               {TIN} \cite{li2019transferable} &  & 17.22 & 13.51 & 18.32\\
               % RP$_{r^2}$C$_{D}$ \cite{li2019cvpr}  & & 17.22 & \textbf{13.51} & 18.32\\
                GCN-HOI~\cite{xu2019cvpr} &  & 14.70& 13.26 & 15.13 \\
            GPNN \cite{qi2018eccv} &  & 13.11  & 9.34  & 14.23 \\
            ICAN \cite{gao2018bmvc} & & 12.80 & 8.53 & 14.07 \\
            % InteractNet \cite{gkioxari2018detecting} & & 9.94         & 7.16          & 10.77 \\       
            % HO-RCNN \cite{chao2018wacv} &    & 7.81          & 5.37          & 8.54 \\ 
             \hline 
    \end{tabular}
     \vspace{-1mm}
    \caption{Comparison with the state-of-the-art methods on HICO-DET test set (mAP).}
    \label{tab:sota-res}
    \vspace{-4mm}
\end{table}

\subsection{More settings} 
\para{Unlabeled data} can also be added into the whole framework: we first obtain all possible human-object pairs in an unlabeled image (US) from the detection result. Given the trained model of the mix-supervised pipeline, we can estimate the marginal HOI class probability for every human-object pair in the unlabeled image. If the probability is larger than a threshold (\eg 0.5), we take the predicted HOI class as the pseudo ground truth for this human-object pair and add it into the network training in the next cycle. The loss function in (\ref{eq:loss}) now includes another term for the unlabeled data, which is formulated similarly to the fully-labeled data with pseudo ground truth. The loss weights among the three terms remain 1.  This process iterates for several cycles until the convergence of the network. 
%two vectors ($p^{ws}_{ij}$ and $p^{fs}_{ij}$) of class probabilities from the weakly- and fully-supervised pipeline, respectively. We apply a weighted sum to $p^{ws}_{ij}$ and $p^{fs}_{ij}$ such as the predicted class label $j$ is obtained via 
%\begin{equation}\label{eq:unlabel}
%    j^* = \argmax_{j}(\lambda p^{ws}_{ij} + (1-\lambda)p^{fs}_{ij}),  p_{ij^*} = \max_j(\lambda p^{ws}_{ij} + (1-\lambda)p^{fs}_{ij})
%\end{equation}
%with an associated probability $p_{ij^*}$. $\lambda$ is a parameter to control the weight between $p^{ws}_{ij}$ and $p^{fs}_{ij}$. It is set to \miaojing{xx}.  

Table~\ref{tab:unlabel} shows the result of WS/FS/US being 30/40/30, where 30\%, 40\% and 30\% percent data are respectively weakly-, fully- and un-supervised (see Sec.~\ref{Sec:experimentsetup}). Results using unlabeled data improves performance when compared to using only WS/FS with 30/40 ratio.
%especially on the non-rare classes which are better detected. This is an interesting trend considering the generally large amount of unlabeled data available.
%Some more settings are evaluated such as WS/US = 30/70 and FS/US = 30/70 in Table~\ref{tab:unlabel}:
%\fix{meaning? -> unlabeled data can be also combined with weakly- or fully-labeled data only under our framework.}   

%Pseudo labels and Unsupervised, results for Full, Rare and Non-Rare	is 15.82, 11.47, 17.11 respectively for 30 FS and pseudo labels and 17.075, 11.178,	18.837 Baseline with 100 FS.

\begin{table}[t]
  	\setlength{\tabcolsep}{3.5pt}
    \centering
    \small
    \begin{tabular}{ccccc} 
    \toprule

   &   \textbf{WS/FS/US} &  \textbf{Full} & \textbf{Rare} & \textbf{Non-Rare} \\
        \midrule
 %       MX-HOI & 30/70/0  & 	&		&	\\
        MX-HOI &  30/40/0  & 16.08 & 12.05 & 17.29 \\ 
          MX-HOI & 30/40/30  & 16.53 & 11.63 & 17.79 \\
      %  MX-HOI &  30/70/0  & 	17.47	&	12.79	& 18.80	\\ 
%         MX-HOI & 0/30/70  & 	&		&	\\
     \bottomrule
    \end{tabular}
    \vspace{-2mm}
 \caption{Adding unlabeled data (US) into MX-HOI. }
    \label{tab:unlabel}
\vspace{-3mm}
\end{table}

\vspace{2mm}
\para{Class-split:} Instead of randomly splitting the dataset images for weak and full supervision, we can randomly split the whole HOI classes into 50\% vs. 50\%. Images from the first 50\% classes are trained with full supervision, the second 50\% with weak supervision. If we train two models separately on the two sets, we got mAP 13.3 and 11.6 on the test set of each own part of classes, respectively. If we train one model over the two sets jointly using MX-HOI, the mAP increases to 14.8 and 13.10. Despite the two sets are of different classes, training them together with more data benefit the performance of both in our pipeline.}

\section{Conclusion}
We present a mixed-supervised HOI detection framework (MX-HOI) which employs two state-of-the-art fully- and weakly-supervised  pipelines. Within this framework, we first introduce a momentum-independent strategy to tackle the adversarial effect of full and weak supervision by separating their gradient history in momentum learning. Second, we introduce an HOI element swapping strategy to harvest hard negatives across images for weakly-labeled data. Unlabeled data can also be leveraged using a “pseudo label” solution where class labels on HOI pairs are provided by the trained mixed-supervised pipeline.  Extensive experiments on the large-scale HICO-DET dataset show that,  with only  30\% fully-labeled data and 70\% weakly-labeled data, our MX-HOI is able to retain 93.3\% accuracy of the setting of 100\% fully-labeled data. Future work will be focused on developing a stronger weakly-supervised HOI detection pipeline to integrate it into our MX-HOI framework. \\

\medskip 

\noindent {\textbf{Acknowledgements:} %This work was done part of READ-IT project funded by the Joint Programming Initiative for Cultural Heritage. 
This work was partially supported by the READ-IT project, funded by the JPI Cultural Heritage   under   the   European   Union   Horizon   2020 R\&I   program   (grant   agreement   No. 699523).
Miaojing Shi was supported by the National Natural Science Foundation of China (NSFC) under Grant No. 61828602. }
%(Reading Europe Advanced Data Investigation Tool) 

\pagebreak
%%%%%%%%% ABSTRACT
% \begin{abstract}
%   The ABSTRACT is to be in fully-justified italicized text, at the top
%   of the left-hand column, below the author and affiliation
%   information. Use the word ``Abstract'' as the title, in 12-point
%   Times, boldface type, centered relative to the column, initially
%   capitalized. The abstract is to be in 10-point, single-spaced type.
%   Leave two blank lines after the Abstract, then begin the main text.
%   Look at previous WACV abstracts to get a feel for style and length.
% \end{abstract}

%%%%%%%%% BODY TEXT

{\small
\bibliographystyle{ieee_fullname}
\bibliography{egbib}

\begin{thebibliography}{10}\itemsep=-1pt

\bibitem{atzmon2016arxiv}
Yuval Atzmon, Jonathan Berant, Vahid Kezami, Amir Globerson, and Gal Chechik.
\newblock Learning to generalize to new compositions in image understanding.
\newblock {\em arXiv preprint arXiv:1608.07639}, 2016.

\bibitem{bansal2020aaai}
Ankan Bansal, Sai~Saketh Rambhatla, Abhinav Shrivastava, and Rama Chellappa.
\newblock Detecting human-object interactions via functional generalization.
\newblock In {\em AAAI}, 2020.

\bibitem{bilen2016cvpr}
Hakan Bilen and Andrea Vedaldi.
\newblock Weakly supervised deep detection networks.
\newblock In {\em CVPR}, 2016.

\bibitem{chao2018wacv}
Yu-Wei Chao, Yunfan Liu, Xieyang Liu, Huayi Zeng, and Jia Deng.
\newblock Learning to detect human-object interactions.
\newblock In {\em WACV}, 2018.

\bibitem{chao2015iccv}
Yu-Wei Chao, Zhan Wang, Yugeng He, Jiaxuan Wang, and Jia Deng.
\newblock Hico: A benchmark for recognizing human-object interactions in
  images.
\newblock In {\em ICCV}, 2015.

\bibitem{cinbis2016pami}
Ramazan~Gokberk Cinbis, Jakob Verbeek, and Cordelia Schmid.
\newblock Weakly supervised object localization with multi-fold multiple
  instance learning.
\newblock {\em IEEE Transactions on Pattern Analysis and Machine Intelligence},
  39(1):189--203, 2016.

\bibitem{cui2018mm}
Zhen Cui, Chunyan Xu, Wenming Zheng, and Jian Yang.
\newblock Context-dependent diffusion network for visual relationship
  detection.
\newblock In {\em ACM MM}, 2018.

\bibitem{delaitre2011nips}
Vincent Delaitre, Josef Sivic, and Ivan Laptev.
\newblock Learning person-object interactions for action recognition in still
  images.
\newblock In {\em NeurIPS}, 2011.

\bibitem{desai2012eccv}
Chaitanya Desai and Deva Ramanan.
\newblock Detecting actions, poses, and objects with relational phraselets.
\newblock In {\em ECCV}, 2012.

\bibitem{desai2010cvprw}
Chaitanya Desai, Deva Ramanan, and Charless Fowlkes.
\newblock Discriminative models for static human-object interactions.
\newblock In {\em CVPR Workshops}, 2010.

\bibitem{deslaers2012pami}
Thomas Deselaers, Alexe Bogdan, and Vittorio Ferrari.
\newblock Weakly supervised localization and learning with generic knowledge.
\newblock {\em International Journal of Computer Vision}, 100:275--293, 2012.

\bibitem{gao2020drg}
Chen Gao, Jiarui Xu, Yuliang Zou, and Jia-Bin Huang.
\newblock Drg: Dual relation graph for human-object interaction detection.
\newblock In {\em ECCV}, 2020.

\bibitem{gao2018bmvc}
Chen Gao, Yuliang Zou, and Jia-Bin Huang.
\newblock ican: Instance-centric attention network for human-object interaction
  detection.
\newblock In {\em BMVC}, 2018.

\bibitem{gkioxari2018detecting}
Georgia Gkioxari, Ross Girshick, Piotr Doll{\'a}r, and Kaiming He.
\newblock Detecting and recognizing human-object interactions.
\newblock In {\em CVPR}, 2018.

\bibitem{gould2008ijcv}
Stephen Gould, Jim Rodgers, David Cohen, Gal Elidan, and Daphne Koller.
\newblock Multi-class segmentation with relative location prior.
\newblock {\em International Journal of Computer Vision}, 80(3):300--316, 2008.

\bibitem{gupta2008eccv}
Abhinav Gupta and Larry~S Davis.
\newblock Beyond nouns: Exploiting prepositions and comparative adjectives for
  learning visual classifiers.
\newblock In {\em ECCV}, 2008.

\bibitem{gupta2009pami}
Abhinav Gupta, Aniruddha Kembhavi, and Larry~S Davis.
\newblock Observing human-object interactions: Using spatial and functional
  compatibility for recognition.
\newblock {\em IEEE Transactions on Pattern Analysis and Machine Intelligence},
  31(10):1775--1789, 2009.

\bibitem{gupta2015arxiv}
Saurabh Gupta and Jitendra Malik.
\newblock Visual semantic role labeling.
\newblock {\em arXiv preprint arXiv:1505.04474}, 2015.

\bibitem{gupta2019iccv}
Tanmay Gupta, Alexander Schwing, and Derek Hoiem.
\newblock No-frills human-object interaction detection: Factorization, layout
  encodings, and training techniques.
\newblock In {\em ICCV}, 2019.

\bibitem{han2018mm}
Chaojun Han, Fumin Shen, Li Liu, Yang Yang, and Heng~Tao Shen.
\newblock Visual spatial attention network for relationship detection.
\newblock In {\em ACM MM}, 2018.

\bibitem{hu2019iccv}
Ronghang Hu, Anna Rohrbach, Trevor Darrell, and Kate Saenko.
\newblock Language-conditioned graph networks for relational reasoning.
\newblock In {\em ICCV}, 2019.

\bibitem{kumar2010cvpr}
M~Pawan Kumar and Daphne Koller.
\newblock Efficiently selecting regions for scene understanding.
\newblock In {\em CVPR}, 2010.

\bibitem{lee2013icmlw}
Dong-Hyun Lee.
\newblock Pseudo-label: The simple and efficient semi-supervised learning
  method for deep neural networks.
\newblock In {\em ICML Workshop}, 2013.

\bibitem{li2017cvpr}
Yikang Li, Wanli Ouyang, Xiaogang Wang, and Xiao'ou Tang.
\newblock Vip-cnn: Visual phrase guided convolutional neural network.
\newblock In {\em CVPR}, 2017.

\bibitem{li2020pastanet}
Yong-Lu Li, Liang Xu, Xinpeng Liu, Xijie Huang, Yue Xu, Shiyi Wang, Hao-Shu
  Fang, Ze Ma, Mingyang Chen, and Cewu Lu.
\newblock Pastanet: Toward human activity knowledge engine.
\newblock In {\em CVPR}, 2020.

\bibitem{li2019transferable}
Yong-Lu Li, Siyuan Zhou, Xijie Huang, Liang Xu, Ze Ma, Hao-Shu Fang, Yanfeng
  Wang, and Cewu Lu.
\newblock Transferable interactiveness knowledge for human-object interaction
  detection.
\newblock In {\em CVPR}, 2019.

\bibitem{liang2017cvpr}
Xiaodan Liang, Lisa Lee, and Eric~P Xing.
\newblock Deep variation-structured reinforcement learning for visual
  relationship and attribute detection.
\newblock In {\em CVPR}, 2017.

\bibitem{mscoco14}
Tsung-Yi Lin, Michael Maire, Serge Belongie, James Hays, Pietro Perona, Deva
  Ramanan, Piotr Doll{\'a}r, and C.~Lawrence Zitnick.
\newblock Microsoft coco: Common objects in context.
\newblock In {\em ECCV}, 2014.

\bibitem{liuamplifying}
Y Liu, Q Chen, and A Zisserman.
\newblock Amplifying key cues for human-object-interaction detection.
\newblock {\em ECCV}, 2020.

\bibitem{liu2019cvpr}
Yuting Liu, Miaojing Shi, Qijun Zhao, and Xiaofang Wang.
\newblock Point in, box out: Beyond counting persons in crowds.
\newblock In {\em CVPR}, 2019.

\bibitem{lu2016eccv}
Cewu Lu, Ranjay Krishna, Michael Bernstein, and Li Fei-Fei.
\newblock Visual relationship detection with language priors.
\newblock In {\em ECCV}, pages 852--869, 2016.

\bibitem{papandreou2015iccv}
George Papandreou, Liang-Chieh Chen, Kevin~P Murphy, and Alan~L Yuille.
\newblock Weakly-and semi-supervised learning of a deep convolutional network
  for semantic image segmentation.
\newblock In {\em ICCV}, 2015.

\bibitem{peyre2019iccv}
Julia Peyre, Ivan Laptev, Cordelia Schmid, and Josef Sivic.
\newblock Detecting unseen visual relations using analogies.
\newblock In {\em ICCV}, 2019.

\bibitem{peyre2017iccv}
Julia Peyre, Josef Sivic, Ivan Laptev, and Cordelia Schmid.
\newblock Weakly-supervised learning of visual relations.
\newblock In {\em ICCV}, 2017.

\bibitem{prest2011pami}
Alessandro Prest, Cordelia Schmid, and Vittorio Ferrari.
\newblock Weakly supervised learning of interactions between humans and
  objects.
\newblock {\em IEEE Transactions on Pattern Analysis and Machine Intelligence},
  34(3):601--614, 2011.

\bibitem{qi2018eccv}
Siyuan Qi, Wenguan Wang, Baoxiong Jia, Jianbing Shen, and Song-Chun Zhu.
\newblock Learning human-object interactions by graph parsing neural networks.
\newblock In {\em ECCV}, 2018.

\bibitem{ramanathan2015cvpr}
Vignesh Ramanathan, Congcong Li, Jia Deng, Wei Han, Zhen Li, Kunlong Gu, Yang
  Song, Samy Bengio, Charles Rosenberg, and Li Fei-Fei.
\newblock Learning semantic relationships for better action retrieval in
  images.
\newblock In {\em CVPR}, 2015.

\bibitem{sadeghi2011cvpr}
Mohammad~Amin Sadeghi and Ali Farhadi.
\newblock Recognition using visual phrases.
\newblock In {\em CVPR}, 2011.

\bibitem{shi2017iccv}
Miaojing Shi, Holger Caesar, and Vittorio Ferrari.
\newblock Weakly supervised object localization using things and stuff
  transfer.
\newblock In {\em ICCV}, 2017.

\bibitem{shi2016eccv}
Miaojing Shi and Vittorio Ferrari.
\newblock Weakly supervised object localization using size estimates.
\newblock In {\em ECCV}, 2016.

\bibitem{shi2015bmvc}
Zhiyuan Shi, Parthipan~Siva Siva, and Tao Xiang.
\newblock Transfer learning by ranking for weakly supervised object annotation.
\newblock In {\em BMVC}, 2015.

\bibitem{sun2019mm}
Xu Sun, Yuan Zi, Tongwei Ren, Jinhui Tang, and Gangshan Wu.
\newblock Hierarchical visual relationship detection.
\newblock In {\em ACM MM}, 2019.

\bibitem{tang2018pami}
Peng Tang, Xinggang Wang, Song Bai, Wei Shen, Xiang Bai, Wenyu Liu, and Alan
  Yuille.
\newblock Pcl: Proposal cluster learning for weakly supervised object
  detection.
\newblock {\em IEEE Transactions on Pattern Analysis and Machine Intelligence},
  42(1):176--191, 2018.

\bibitem{tang2017cvpr}
Peng Tang, Xinggang Wang, Xiang Bai, and Wenyu Liu.
\newblock Multiple instance detection network with online instance classifier
  refinement.
\newblock In {\em CVPR}, 2017.

\bibitem{ulutan2020vsgnet}
Oytun Ulutan, ASM Iftekhar, and Bangalore~S Manjunath.
\newblock Vsgnet: Spatial attention network for detecting human object
  interactions using graph convolutions.
\newblock In {\em CVPR}, 2020.

\bibitem{wan2019pose}
Bo Wan, Desen Zhou, Yongfei Liu, Rongjie Li, and Xuming He.
\newblock Pose-aware multi-level feature network for human object interaction
  detection.
\newblock In {\em ICCV}, 2019.

\bibitem{wang2020learning}
Tiancai Wang, Tong Yang, Martin Danelljan, Fahad~Shahbaz Khan, Xiangyu Zhang,
  and Jian Sun.
\newblock Learning human-object interaction detection using interaction points.
\newblock In {\em CVPR}, 2020.

\bibitem{xu2019cvpr}
Bingjie Xu, Yongkang Wong, Junnan Li, Qi Zhao, and Mohan~S Kankanhalli.
\newblock Learning to detect human-object interactions with knowledge.
\newblock In {\em CVPR}, 2019.

\bibitem{xu2019mm}
Wanru Xu, Jian Yu, Zhenjiang Miao, Lili Wan, and Qiang Ji.
\newblock Prediction-cgan: Human action prediction with conditional generative
  adversarial networks.
\newblock In {\em ACM MM}, 2019.

\bibitem{yang2020nips}
Yukuan Yang, Fangyu Wei, Miaojing Shi, and Guoqi Li.
\newblock Restoring negative information in few-shot object detection.
\newblock In {\em NeurIPS}, 2020.

\bibitem{yang2019arxiv}
Zhaohui Yang, Miaojing Shi, Yannis Avrithis, Chao Xu, and Vittorio Ferrari.
\newblock Training object detectors from few weakly-labeled and many unlabeled
  images.
\newblock {\em arXiv preprint arXiv:1912.00384}, 2019.

\bibitem{yatskar2016cvpr}
Mark Yatskar, Luke Zettlemoyer, and Ali Farhadi.
\newblock Situation recognition: Visual semantic role labeling for image
  understanding.
\newblock In {\em CVPR}, 2016.

\bibitem{yu2017iccv}
Ruichi Yu, Ang Li, Vlad~I Morariu, and Larry~S Davis.
\newblock Visual relationship detection with internal and external linguistic
  knowledge distillation.
\newblock In {\em ICCV}, 2017.

\bibitem{zhang2017iccv}
Hanwang Zhang, Zawlin Kyaw, Jinyang Yu, and Shih-Fu Chang.
\newblock {PPR-FCN:}: Weakly supervised visual relation detection via parallel
  pairwise {R-FCN}.
\newblock In {\em ICCV}, 2017.

\bibitem{zheng2019mm}
Sipeng Zheng, Shizhe Chen, and Qin Jin.
\newblock Visual relation detection with multi-level attention.
\newblock In {\em ACM MM}, 2019.

\bibitem{zhou2019mm}
Hao Zhou, Chongyang Zhang, and Chuanping Hu.
\newblock Visual relationship detection with relative location mining.
\newblock In {\em ACM MM}, 2019.

\bibitem{zhou2019iccv}
Penghao Zhou and Mingmin Chi.
\newblock Relation parsing neural network for human-object interaction
  detection.
\newblock In {\em ICCV}, 2019.

\end{thebibliography}
}

\end{document}